\documentclass[runningheads]{llncs}
\usepackage{graphicx}
\usepackage{amsmath}
\usepackage{multirow}
\usepackage{subcaption}
\usepackage[font=scriptsize]{caption}
\newcommand{\ignore}[1]{}
\def\sym#1{\ifmmode^{#1}\elsel\(^{#1}\)\fi}

\begin{document}
%

\title{Using Adaptive Experiments to Rapidly Help Students}

%
\titlerunning{Exploring the Use of Adaptive Experiments}
%
\author{Angela Zavaleta-Bernuy\inst{1} \and
Qi Yin Zheng\inst{1} \and
Hammad Shaikh\inst{1} \and Jacob Nogas\inst{1} \and Anna Rafferty\inst{2} \and Andrew Petersen\inst{1} \and Joseph Jay Williams\inst{1}}
\authorrunning{Zavaleta-Bernuy et al.}
%
\institute{University of Toronto \and Carleton College}
%
\maketitle              
%

\begin{abstract}
Adaptive experiments can increase the chance that current students obtain better outcomes from a field experiment of an instructional intervention. In such experiments, the probability of assigning students to conditions changes while more data is being collected, so students can be assigned to interventions that are likely to perform better. Digital educational environments lower the barrier to conducting such adaptive experiments, but they are rarely applied in education. One reason might be that researchers have access to few real-world case studies that illustrate the advantages and disadvantages of these experiments in a specific context. We evaluate the effect of homework email reminders in students by conducting an adaptive experiment using the Thompson Sampling algorithm and compare it to a traditional uniform random experiment. We present this as a case study on how to conduct such experiments, and we raise a range of open questions about the conditions under which adaptive randomized experiments may be more or less useful.

\end{abstract}
\section{Introduction}

Instructors frequently look for ways to support their students and improve their performance. With access to online learning environments, instructors can gather feedback in a larger scale setting. Since optimal instructional designs and scaffolds may not be known ahead of time, multiple possibilities can be tested using Uniform Random (UR) A/B experiments, also known as randomized control trials. In a UR A/B experiment, students are uniformly assigned to the different conditions that an instructor or researcher would like to test to learn about their relative effectiveness.
One challenge of this approach is how to use data more rapidly to help current students. To mitigate this, we can aim to maximize total learning by having most students being subject to the more effective conditions as they become known.



%

Adaptive randomization is an effective strategy for assigning more students to the current most optimal condition, while retaining the ability to test other conditions. We use a Multi-Armed Bandit (MAB) algorithm that uses machine learning to increase the number of students assigned to the current most effective condition (or \textit{arm}) \cite{lomas2016interface}, \cite{williams2018enhancing}. MAB are commonly used for rapid use of data in different areas such as marketing to optimize the benefits of the users and balance exploration vs. exploitation \cite{lomas2016interface}, \cite{rafferty2018bandit}. For this study, we used Thompson Sampling (TS), a probability matching algorithm, where the probability of assignment is proportional to the probability that the arm is optimal \cite{lomas2016interface}. 

In this paper, we present a real-world experiment to illustrate the benefits and limitations of using UR A/B experiments and TS in educational settings. First, we use UR A/B experiments to evaluate different versions of emails in a homework reminder intervention to determine if a more effective version can be identified. We then compare the results of UR A/B experiments with the TS results to study its performance and benefits. Our experimental design allows us to compare classical balanced A/B comparisons side-by-side with a TS adaptive experiment to evaluate the trade-offs of using each of these methods.


\section{Multi-Armed Bandit (MAB) Algorithms}

To optimize the experience of students, we use the TS algorithm designed to solve MAB problems, useful for adaptively assigning participants to conditions.
 \cite{williams2018enhancing}.

The stochastic MAB problem is the problem of sequentially choosing from a discrete set of actions to maximize cumulative reward, where a reward is some measure of the effectiveness of the chosen action (arm). In this paper, we focus on the MAB problem with binary rewards. More precisely, we choose between $K$ versions (arms), and we denote the choice of action at step $t$ of the experiment by $x_t$. Assuming we choose the $k-{th}$ arm, where $k \in \{1, 2, \ldots, K\}$, then $x_t = k$, and we receive a reward $r_t$ with probability $p_k$. 

TS shows strong empirical performance in maximizing the cumulative reward \cite{sutton2018reinforcement}. TS is a Bayesian algorithm that maintains a posterior distribution over each reward $p_k$. In our case, we use a Beta prior with parameters $\alpha_k$ and $\beta_k$. Arms are chosen by sampling values from the posteriors over each arm, and choosing the arm corresponding to the highest sample drawn. The posterior distribution is then updated based on the chosen action $x_t=k$ and observed reward $r_t$. We use a uniform prior for each arm, $\alpha_k=\beta_k=1$, for all $k \in \{1, 2 ,..., K\}$. 

\[(\alpha_k, \beta_k) \leftarrow 
\begin{cases}
(\alpha_k, \beta_k) & \text{if } x_t \neq k\\
(\alpha_k, \beta_k) + (r_t, 1 - r_t) & \text{if } x_t = k\\
\end{cases}\]

\section{Methods: Traditional and Adaptive Experimentation}


For three consecutive weeks, we tested four different versions of the emails to investigate which might be more effective in leading students to click on the homework link appended in the email. To evaluate how TS adapts the distribution of students to each email version, we sent the messages in four different batches on consecutive days of the week (Tuesday to Friday). 

Using UR, each of the four email reminders has the same probability of being assigned to a student. For the TS algorithm, the probability of assignment of the email reminders version is proportional to the \textit{reward} (in our case, click rate) in all the previous batches, which is updated after each batch.

For our interventions, we used two different variations of the TS algorithm. For Weeks 1 and 2, we used a UR-TS Hybrid where the TS updating of the probability that an arm has the highest click rate used data from the UR participants too. This hybrid is called $\epsilon$[0.5]-TS, because with epsilon (epsilon = 0.5) probability, arms are assigned using UR. This takes inspiration from past algorithms like epsilon-greedy \cite{watkins1989learning} and top-two TS \cite{russo2016simple}. This is interesting to investigate because scientists may want to get the benefits of obtaining data under UR (in case TS introduces biases \cite{rafferty2019statistical}) for analysis, while also then using that data to \textit{improve} the performance of the TS algorithm. For Week 3, we applied traditional TS that did not use the data from the UR assignment.

\section{Analysis and Results}

Using a panel regression with week fixed effects (i.e., include indicators for Week2-$\epsilon$[0.5]-TS and Week3-TS), we can aggregate the uniform portion of the experiment for the three weeks to evaluate the effects of the four arms on student click rate. We find that the average click rate in our sample across the three weeks is around $19\%.$ All four arms had click rates are within 2\% of each other and their difference is not statistically significant regardless of time and participant effects. The lack of an optimal arm suggests that all students were assigned to fairly similar treatments. These results are robust to also including student fixed effects within the panel regression model.  




For both of the later weeks, $\epsilon$[0.5]-TS behaves similarly, favouring one particular arm. Some arms are assigned a substantially higher number of students and the others far fewer, but the seemingly favoured arm varies across the two weeks and is not consistent in Week2-$\epsilon$[0.5]-TS. In Table \ref{tab:ts_reward}, we show the cumulative reward per arm and per batch, which influenced the probability of assignment. Even though the probability of assignment aligns with the click rates from the previous batch, the algorithm drastically shifts to the arm with the highest reward and leaves minimal room for exploration in batch 3. As the click rates were not consistent across the different batches, we are unable to identify the presence of the most efficient arm, which is reflected on the cumulative rewards in Table \ref{tab:ts_reward}, and could be a result of there being minimal differences between arms.





\begin{table}[!t]
    \centering
    \begin{tabular}{c c|c c c c c c c c}
 &  & \multicolumn{2}{c}{Arm 1}  & \multicolumn{2}{c}{Arm 2} & \multicolumn{2}{c}{Arm 3}  & \multicolumn{2}{c}{Arm 4} \\
 &  &  CCR &  PA &   CCR &  PA &  CCR &  PA & CCR &  PA\\
 \hline
\multirow{4}{*}{\shortstack{Week1-$\epsilon$[0.5]-TS}} & Batch 1 & 0.200 & 0.117 & 0.277 & \textbf{0.659} & 0.212 & 0.177 & 0.167 & 0.047\\
 & Batch 2 & 0.219 & \textbf{0.466} & 0.22 & 0.443 & 0.149 & 0.040 & 0.152 & 0.052\\
 & Batch 3 & 0.206 & 0.434 & 0.209 & \textbf{0.452} & 0.137 & 0.017 & 0.163 & 0.097\\
 & Batch 4 & 0.163 & 0.077 & 0.205 & \textbf{0.697} & 0.161 & 0.104 & 0.162 & 0.122\\
 \hline
\multirow{4}{*}{\shortstack{Week2-$\epsilon$[0.5]-TS}} & Batch 1 & 0.213 & 0.116 & 0.086 & 0.000 & 0.246 & 0.225 & 0.300 & \textbf{0.659}\\
 & Batch 2 & 0.198 & 0.082 & 0.14 & 0.004 & 0.244 & 0.311 & 0.266 & \textbf{0.603}\\
 & Batch 3 & 0.197 & 0.065 & 0.186 & 0.041 & 0.261 & \textbf{0.666} & 0.235 & 0.229\\
 & Batch 4 & 0.194 & 0.052 & 0.209 & 0.109 & 0.249 & \textbf{0.505} & 0.240 & 0.334\\
 \hline
\multirow{4}{*}{\shortstack{Week3-TS}} & Batch 1 & 0.231 & \textbf{0.477} & 0.153 & 0.056 & 0.22 & 0.389 & 0.157 & 0.078\\
 & Batch 2 & 0.233 & \textbf{0.926} & 0.137 & 0.030 & 0.135 & 0.033 & 0.120 & 0.011\\
 & Batch 3 & 0.183 & \textbf{0.510} & 0.129 & 0.039 & 0.133 & 0.070 & 0.174 & 0.381\\
 & Batch 4 & 0.181 & 0.405 & 0.152 & 0.086 & 0.144 & 0.076 & 0.181 & \textbf{0.433}\\
    \end{tabular}
    \caption{The Table presents the cumulative click rate (CCR) and the probability of assignment (PA) for each arm in the next batch for the three weeks of interventions. The arms represent a version of the message. Each batch represents a separate day of message deployments: Monday (Batch 1), Tuesday (Batch 2), Wednesday (Batch 3), and Thursday (Batch 4). The cumulative click rate of each arm represents the percentage of people who received the message that clicked accumulated up to and including that batch. The probability of assignment is the probability that a particular arm will be chosen in the next batch calculated using 1000000 Monte Carlo simulations. Bold values highlight the arm with the highest probability of assignment for each batch on every week. }
    \label{tab:ts_reward}
\end{table}

\ignore{
\begin{figure}[!ht]
    \centering
    \includegraphics[scale=0.17]{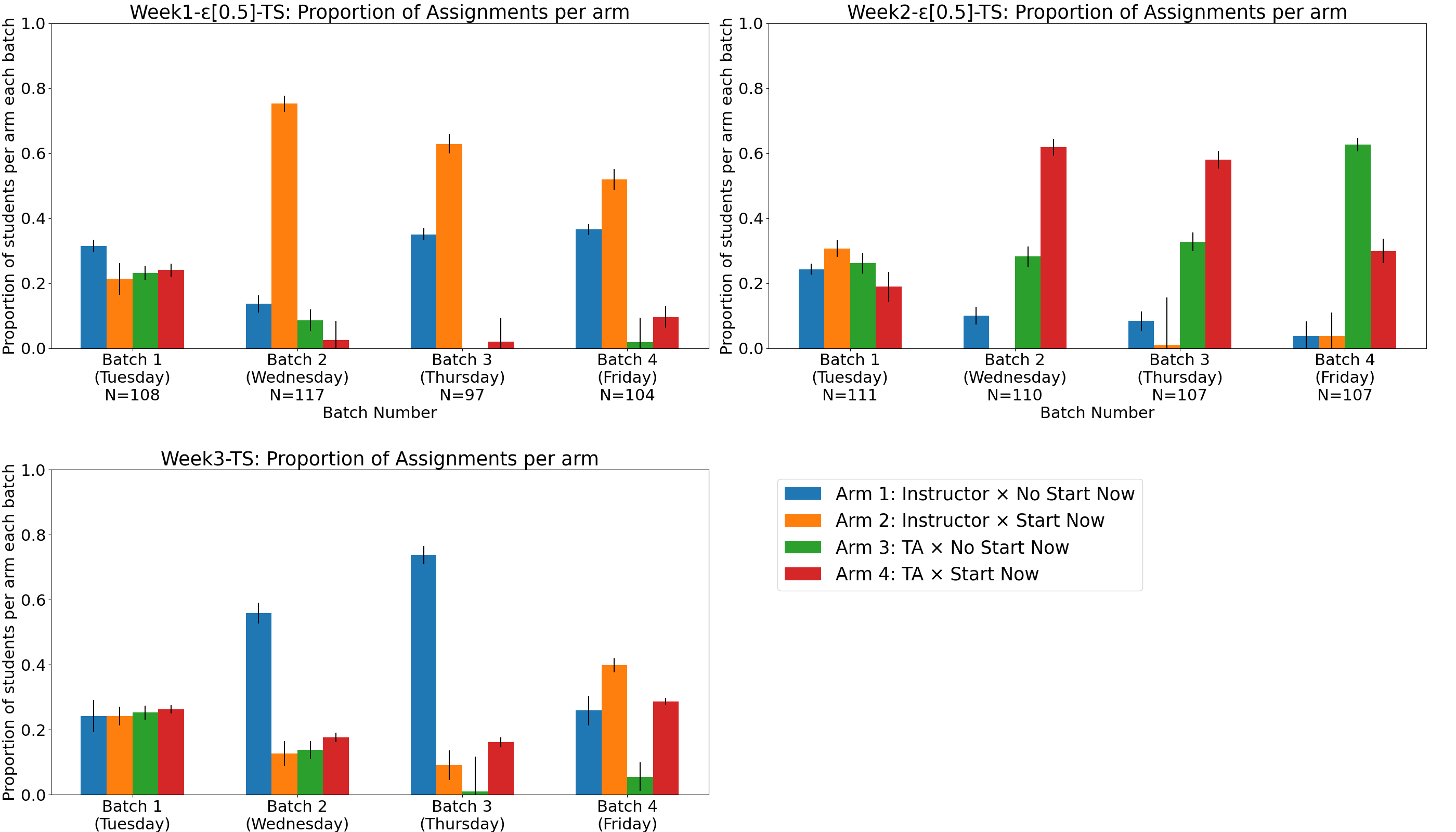}
    \caption{The Figure shows the TS assignment distribution for the three weeks of the intervention. Each arm represent a version of the message. Each batch represents a separate day of message deployments: Tuesday (Batch 1), Wednesday (Batch 2), Thursday (Batch 3), and Friday (Batch 4).}
    \label{fig:ts_assign}
\end{figure}
}

\section{Discussion \& Limitations}

We present a case study of adaptive experimentation, using the Thompson Sampling MAB algorithm, when a difference between arms is not observed (i.e., the arms/conditions are equally effective), according to a traditional Uniform Random experiment. We illustrate how TS may randomly favour an arm, even when giving this arm more frequently has no consequences for participants. The TS algorithm minimizes regret --- it aims to keep participants from being assigned to sub-optimal arms --- so in the case where arms are equivalent, one could argue that any of them could be presented. However, this can be problematic for scientific inference and statistical analysis \cite{rafferty2019statistical}.



One limitation is that there might have been unobserved confounding variables that caused the click rates to change in one particular batch (e.g., students had an assignment due or a test), which will also affect the algorithm – this is one concern to keep in mind in applying MAB algorithms for adaptive experimentation. 

\section{Conclusion \& Future Work}
This paper provides an example of a real-world intervention using adaptive experiments, which can help instructors and researchers to use the results from experiments to more rapidly benefit students. We illustrate an instance of conducting such experiments using the TS algorithm, where the results suggest there is no difference between arms/conditions. We hope that this paper provides a first step for instructors and researchers to investigate adaptive experimentation in education. Future work can explore how the algorithm behaves in a wider variety of scenarios, such as different batch sizes and structure, and alternative differences between the arm/conditions.

\section{Acknowledgements}

This work was partially supported by the Natural Sciences and Engineering Research Council of Canada (NSERC) (\#RGPIN-2019-06968), as well as by the Office of Naval Research (ONR) (\#N00014-21-1-2576).
%
%
%
%
%
\bibliographystyle{splncs04}
\bibliography{mybibliography}
\end{document}